\documentclass{article}
\usepackage{main}

\usepackage{times}
\usepackage{soul}
\usepackage{url}
\usepackage[hidelinks]{hyperref}
\usepackage[utf8]{inputenc}
\usepackage[small]{caption}
\usepackage{graphicx}
\usepackage{amsmath}
\usepackage{amsthm}
\usepackage{booktabs}
\usepackage{algorithm}
\usepackage{algorithmic}
\usepackage{amsfonts}
\usepackage{amsmath}

\title{Matching Text with Deep Mutual Information Estimation}

\iftrue
\author{
Xixi Zhou$^1$\and
Chengxi Li$^1$\and
Jiajun Bu$^1$\and
Chengwei Yao$^1$\and
Keyue Shi$^1$\and
Zhi Yu$^1$\and
Zhou Yu$^2$
\affiliations
$^1$Zhejiang University\\
$^2$University of California, Davis\\
\emails
\{xixi.zxx, chengxili, bjj, yaochw, shikeyue, yuzhirenzhe\}@zju.edu.cn,
\{joyu\}@ucdavis.edu
}
\fi

\begin{document}

\maketitle

\begin{abstract}
Text matching is a core natural language processing research problem. How to retain sufficient information on both content and structure information is one important challenge. In this paper, we present a neural approach for general-purpose text matching with deep mutual information estimation incorporated. Our approach, Text matching with Deep Info Max (TIM), is integrated with a procedure of unsupervised learning of representations by maximizing the mutual information between text matching neural network's input and output. We use both global and local mutual information to learn text representations. We evaluate our text matching approach on several tasks including natural language inference, paraphrase identification, and answer selection. Compared to the state-of-the-art approaches, the experiments show that our method integrated with mutual information estimation learns better text representation and achieves better experimental results of text matching tasks without exploiting pretraining on external data.
\end{abstract}

\section{Introduction}
Text matching is an important research area in several natural language processing (NLP) applications, including, but not limited to information retrieval, natural language inference, question answering and paraphrase identification. In these applications, a model estimates the similarity or relations between two input text sequences and two problems will arise in the process. The first, also common in many NLP tasks, is how to efficiently model or represent texts. The second, specifically for the text matching task, is how to bridge the information gap between two text sequences of non-comparable lengths.

Text matching approaches have successfully introduced many encoder methods or constructed their hybrids to represent texts. Although the representation methods significantly advanced the fields of natural language processing as well as its downstream tasks including text matching applications, they have limitations in transferring information from the inputs to the output representations. Some of them lose important information while handling a fairly long sequence of words, while others that focus on learning the local features are inadequate to represent complex long-form documents. For text matching tasks, it is crucial that the text representations should retain as much useful information of the input data as possible. The other problem of text matching is how to bridge the information gap between two text sequences of lengths with different scales, such as short-short text matching, long-long text matching, and short-long text matching. In all these types, the core information is always hard to be extracted from texts, not only because of the text representation problem above, but also because of different text structures. 

Recently, interests have shifted toward mutual information (MI) maximization of representation across multiple domains, including computer vision and NLP. To efficiently model or represent both sides of text pairs in text matching, a natural idea is to train a representation-learning network to maximize the MI between text inputs and representation outputs before matching. However, MI is difficult to estimate especially in high-dimensional and continuous representation space. Fortunately, the recent theoretical breakthrough has made it possible to effectively compute MI between high dimensional input/output pairs of deep neural networks \cite{Belghazi2018,hjelm2018}. Early attempts have been made to solve NLP tasks like text generation \cite{qian2019} and some other kind of tasks like cross-modal retrieval \cite{wei2019} with MI maximization. 

In this paper, we introduce deep mutual information estimation technique, as known as Deep InfoMax (DIM, \cite{hjelm2018}) into text matching task. We design a deep MI estimation module to maximize the MI between input text pairs and their learned high-level representations. We start with the text matching neural network model of \cite{yang2019simple}, and design a wrapping-mode training architecture. In our architecture, we take the whole text matching network as the encoder while MI between the inputs and the outputs is estimated and maximized so that learned representations can retain information of the input data to a great extent. Moreover, maximizing MI between the input data and the encoder output (global MI) is often insufficient for learning useful representations. Recently the method on maximizing the local MI between the representation and local regions of the input (e.g. patches rather than the complete text) is presented (\cite{hjelm2018}), where the very representation is encouraged to have high MI with all the patches.

So, to preserve the complex structural information and solve the structure difficulty in text matching on varying-length texts, we split input texts into segments as local features, and then maximize the average MI between the high-level representation and local patches of the input text. Our proposed method works effectively and efficiently according to experimental results. The main contributions of this paper are summarized as follows:
\begin{itemize}
\item We propose a deep neural network with deep mutual information estimation to solve problems of text matching. To the best of our knowledge, this work is the first attempt to apply mutual information neural estimation to improving both representation quantity and diversity of text structures in text matching tasks.
\item We integrate the global and local mutual information maximization for texts to help well preserve the information in the process between input and output representation. Our model has fewer parameters and doesn't rely on pretraining on external data, compared to large representation models. This is meaningful to different text matching tasks.
\item Experimental results on four benchmark datasets across four different tasks are all on par with or even above the state-of-the-art methods, which demonstrate the high effectiveness of our method on text matching tasks. 
\end{itemize}

\begin{figure}[H]
  \centering
  \includegraphics[width=90mm]{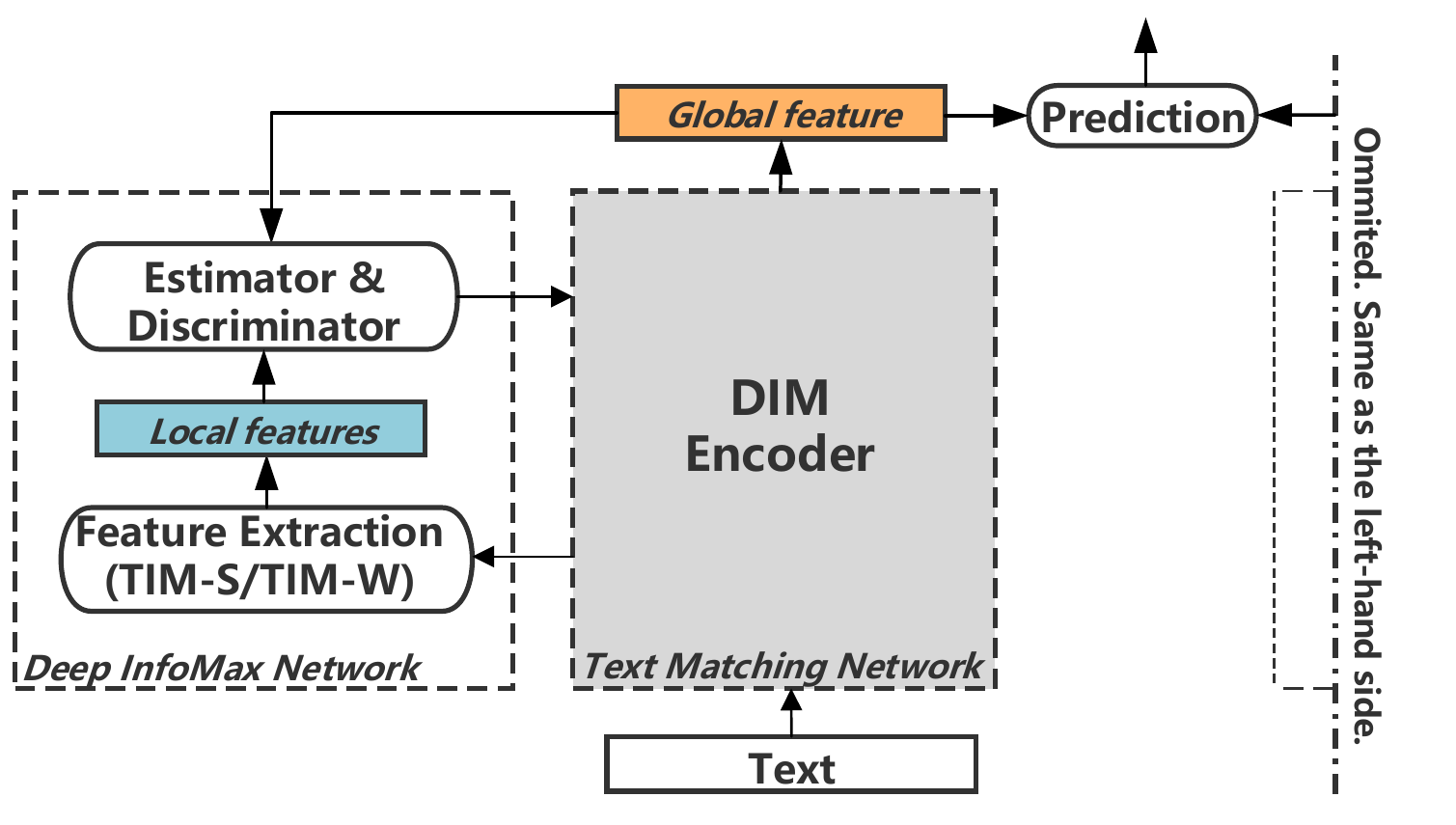}
  \caption{\textbf{Architecture overview of TIM. } The DIM Encoder will be detailed later in the section \ref{sec:dim-encoder} and shown clear in Figure \ref{fig-TIM-RE2}.}
  \label{fig-TIM-overview}
\end{figure}

\section{Related Work}

The first-generation encoder methods of word embeddings, for instance, Word2Vec (\cite{Mikolov2013}) and Doc2Vec (\cite{Le2014}), learn embedding vectors as text representations based on different text structure levels, such as words, sentences, paragraphs and documents. They are introduced into several text matching models in which typical similarity metrics are employed to compute the matching scores of two text vectors (WMD \cite{Kusner2015}). Besides, some latent variable models are introduced into text matching tasks, too. They extract hidden topics from texts, and then the texts can be compared based on their hidden topic representations (\cite{Gong2018}). Recently, deep neural networks become the most popular models for better text representations in NLP tasks, such as convolutional neural networks (CNNs), recurrent neural networks (RNNs) and Long Short-Term Memory architectures (LSTM). Accordingly, many text matching applications take these models as text encoders in their matching processes: \cite{Severyn2015} ranks short text pairs using CNN which preserves local information in the text representation, \cite{Mueller2016} treats texts as a sequence of words in their representation processes and then take RNN as text encoders for text matching on sentences and long-form texts, \cite{tai2015} shows superiority for representing sentence meaning over a sequential LSTM, and \cite{tan2016} introduces LSTM to construct better answer representations in question-answer matching. Nowadays, the state-of-the-art representation methods focus on the contextual token representation - to train an encoder to represent words in their specific context, such as BERT and XLNet.

In text matching tasks, a comparably long text may lose its local information after being encoded as a fixed-sized representation. Some of the previous studies (\cite{tan2016}) exploit attention mechanism to distill important words from sentences, but valuable information can still be diluted within a large number of sentences in long-form texts. On the other hand, representation of a short text has the sparse problem and may lose the global information of word co-occurrency. For this, some previous studies, such as \cite{yang2019simple} typically, employ alignment architecture to rich the mutual information between the sequence pair in matching and introduces augmented residual connections for the encoder for inputs to retain as much information as possible in its outputs. \cite{Liu2018ImprovedTM} focuses on matching question/answer (QA) and adopts the generative adversarial network (GAN) to enhance mutual information by rewriting questions in QA tasks. However MI is able to quantify the dependence of two random variables and to measure non-linear statistical dependencies between variables. \cite{Belghazi2018} implements MI estimation in high-dimensional and continuous scenarios and effectively computed MI between high dimensional input/output pairs of deep neural networks (MINE). \cite{hjelm2018} formalizes Deep InfoMax (DIM), which makes it possible to prioritize global or local information and to tune the suitability of learned representations for classification or reconstruction-style tasks. Inspirited by DIM, we introduce deep mutual information estimation and maximization to our deep neural model for more general text matching tasks.

\section{Methodology}
We adopt the neural architecture on text matching introduced in RE2 \cite{yang2019simple} and apply MI estimation and maximization method to the representation part of the base text matching architecture. We intend to maximize mutual information of texts in the matching process, but if text matching encoders pass information from only some parts of input, this does not increase the MI with any other parts. Based on this, our model introduces DIM to leverage local regions of the input for better text representation, for the same representation is encouraged to have high MI with all patches, and this mechanism will exert influence on all input data shared across patches. Besides, DIM has the representational capacity of deep neural networks. Therefore, it is very suitable for mutual information estimation of high dimensional data including text data.

For the text matching task, our model employs the local DIM framework to estimate and maximize MI. The overall framework of our proposed architecture is presented in Figure \ref{fig-TIM-overview}. In the DIM network on the left hand of Figure \ref{fig-TIM-overview}, multiple feature maps, treated as \textit{local features}, are extracted from one input text by our \textbf{Feature Extraction Method} (section \ref{sec:feature}). The local features reflect some structural aspects of the text data, e.g. spatial locality. For the \textit{global feature}, as shown in the right hand of Figure \ref{fig-TIM-overview}, we take the whole text matching neural network as the \textbf{DIM Encoder} (section \ref{sec:dim-encoder}) of our model, and we take the high-level output representation from its pooling layer as the global feature vector for DIM. Here the DIM network shares the high-level representation with the text matching network output. It is because the base text matching network and MI estimator are optimizing the loss for the same purpose and require similar computations. To implement the DIM model to the base text matching neural network, in the following subsections, we first propose our feature extraction method for text data. Then we describe the base text matching neural network as our DIM encoder. At last, we propose our \textbf{DIM Estimator and Discriminator} (section \ref{sec:dim}) for MI maximization of text matching.

\subsection{Feature Extraction for Varying-Length Text}
\label{sec:feature}
\begin{figure}[H]
    \centering
    \includegraphics[width=85mm]{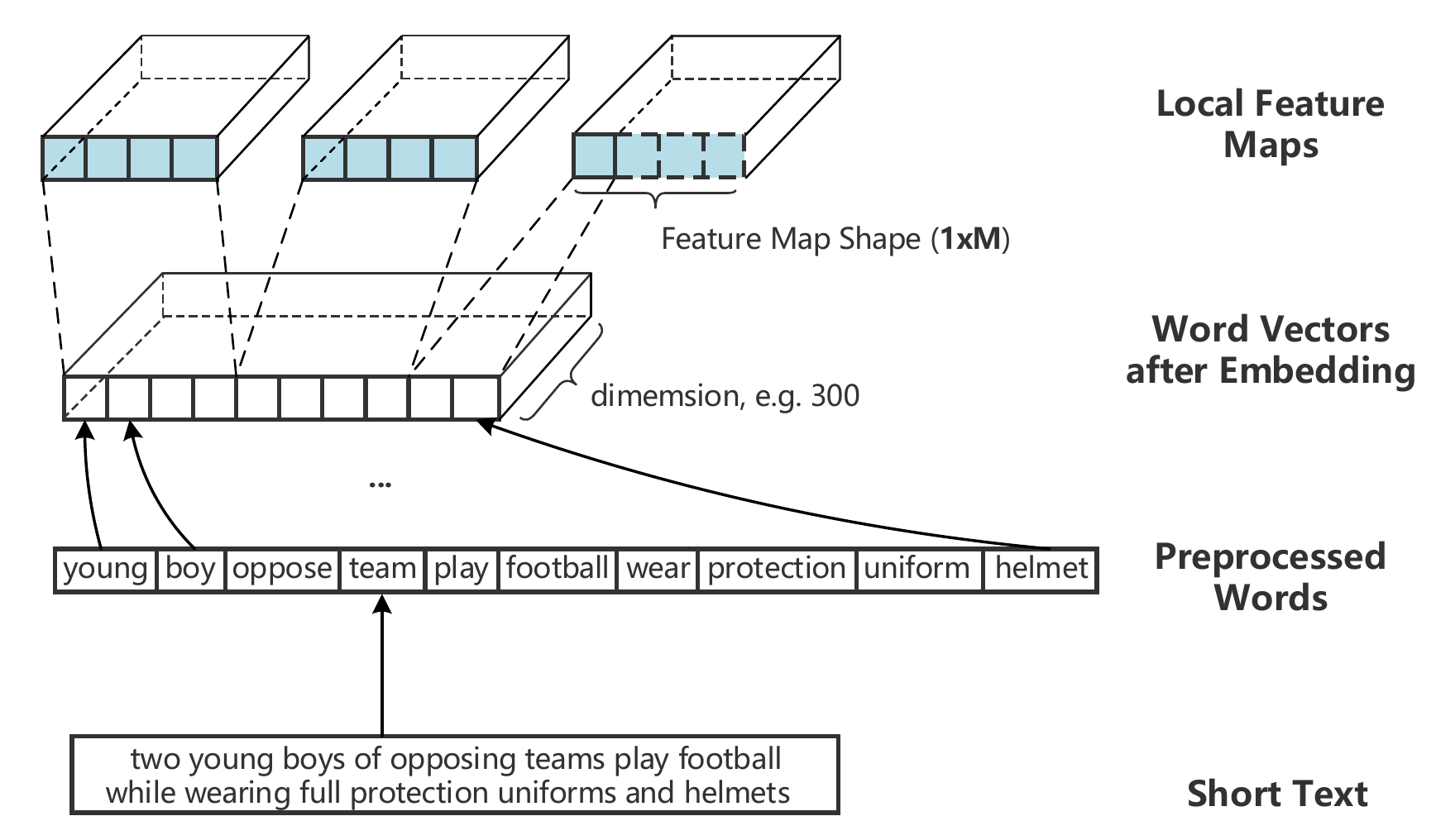}
    \caption{\textbf{Local Feature Extraction of Word Mode (TIM-W).} This mode is for maximizing MI of short texts.}
    \label{fig-TIM-W}
\end{figure}

\begin{figure}[H]
    \centering
    \includegraphics[width=85mm]{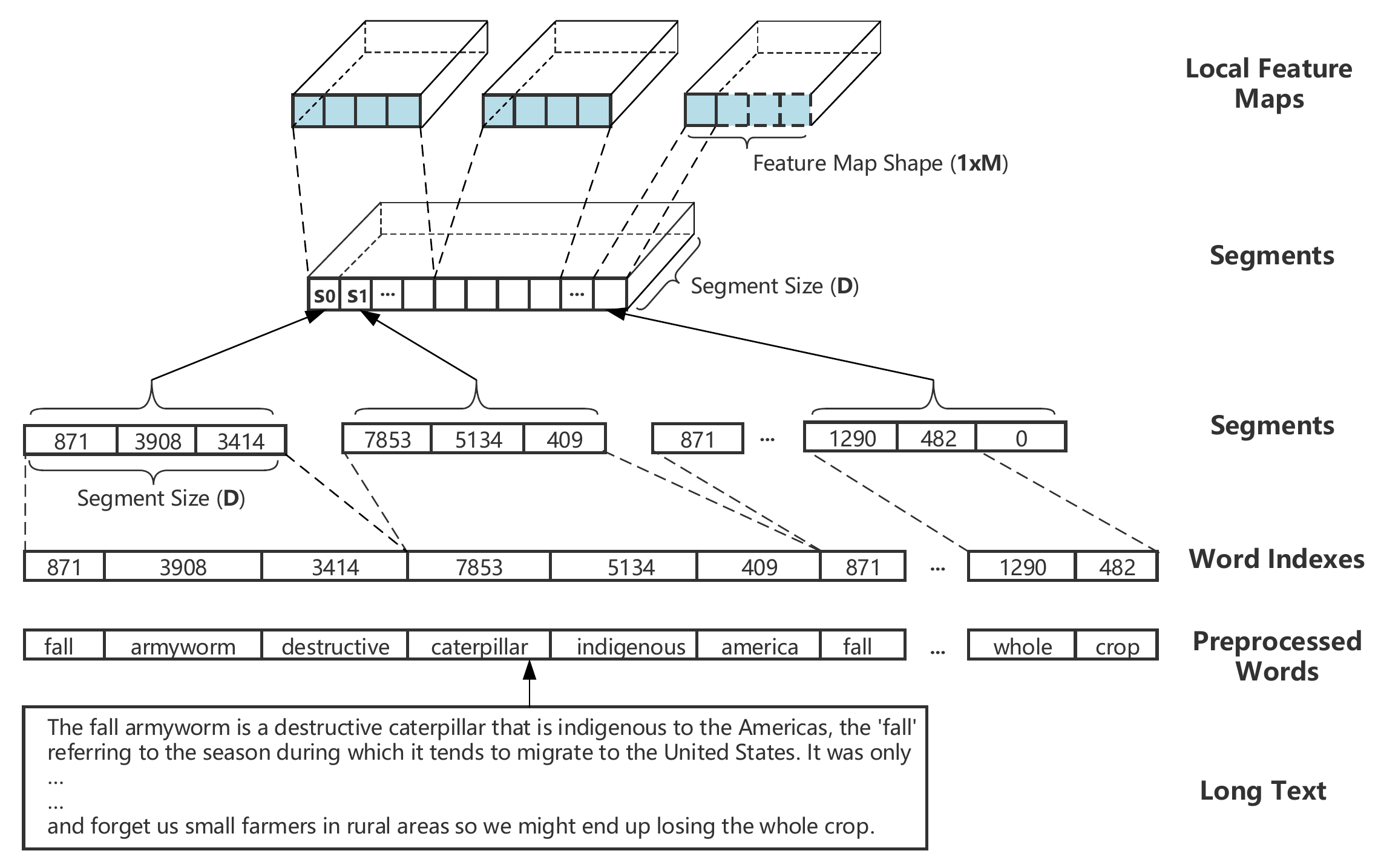}
    \caption{\textbf{Local Feature Extraction of Segment Mode (TIM-S).} This mode is for maximizing MI of long texts.}
    \label{fig-TIM-S}
\end{figure}

First we generate feature maps, $C(X) := {\{C^{(i)}\}}_{i=1}^{1\times{M}}$, for input $X$. In this step, we convert a text to multiple tensors of the same shape, $1\times{M}$, and generate fixed-sized feature maps for using the DIM method. What we need to consider is how to maintain as much useful information of the source text as possible in these feature maps. Therefore, according to different lengths situation of the text pair in the dataset, we propose two generation modes of feature maps separately for short text data and long text data, named word mode (TIM-W, Figure \ref{fig-TIM-W}) and segment mode (TIM-S, Figure \ref{fig-TIM-S}).

The TIM-W is mainly used for short texts to generate feature maps. We observe some universally-used short text datasets, including SNLI, SciTail, Quora and WikiQA, where texts are mostly in the tens of word scale, or are most in the tens of word range. For these cases, we propose the TIM-W to extract feature maps based on words and their embeddings to retain more semantic relevance information in a short text. We convert the short text into a word vector list, denoted as $T$ = ($v_0$, $v_1$,\dots,$v_{n-1}$), in which $v_i$ is a high-dimensional (e.g. 300 dimensions) vector calculated by a simple Word2Vec embeddings. The shape of the feature map ($1$x$M$) is fixed while DIM network is initialized in advance, where $M<n$. The we group the $n$ word vectors into feature maps. We pad the last feature map with zero vectors if the number of the last group of vectors is not big enough to fill all space of the last feature map. The TIM-W mode is shown in Figure \ref{fig-TIM-W}.

For a long text dataset, using a higher-dimensional word embedding to encode a long text will cause high space/time complexity while its texts already have much richer information than short texts. So we propose TIM-S to generate fix-size feature maps for a long text in our text matching model. First, we represent each word of a long text with a word index number defined in a relevant vocabulary: $T$ = ($w_0$, $w_1$,\dots, $w_{n-1}$). Then we divide $T$ into segments with the same fixed length according to the preset segment size ($D$), $S$ = ($s_0$, $s_1$ \dots), where each $s_i$ contains $D$ word indexes and the last segment is padded with zeros at the end of it. Then we group the segments into $M$ feature maps, which shapes are of fixed size, $1$x$M$. The segment size $D$ and feature shape $M$ are set when initializing DIM network in advance. If the last group of segments is not enough to meet the size of the last feature map, we will pad zero-element segments for the last feature map. Finally, the long input text is represented as multiple fix-size local feature maps. The process is shown in Figure \ref{fig-TIM-S}.

\subsection{Text Matching Neural Layers}
\label{sec:dim-encoder}
\begin{figure}[H]
  \centering
  \includegraphics[height=88mm]{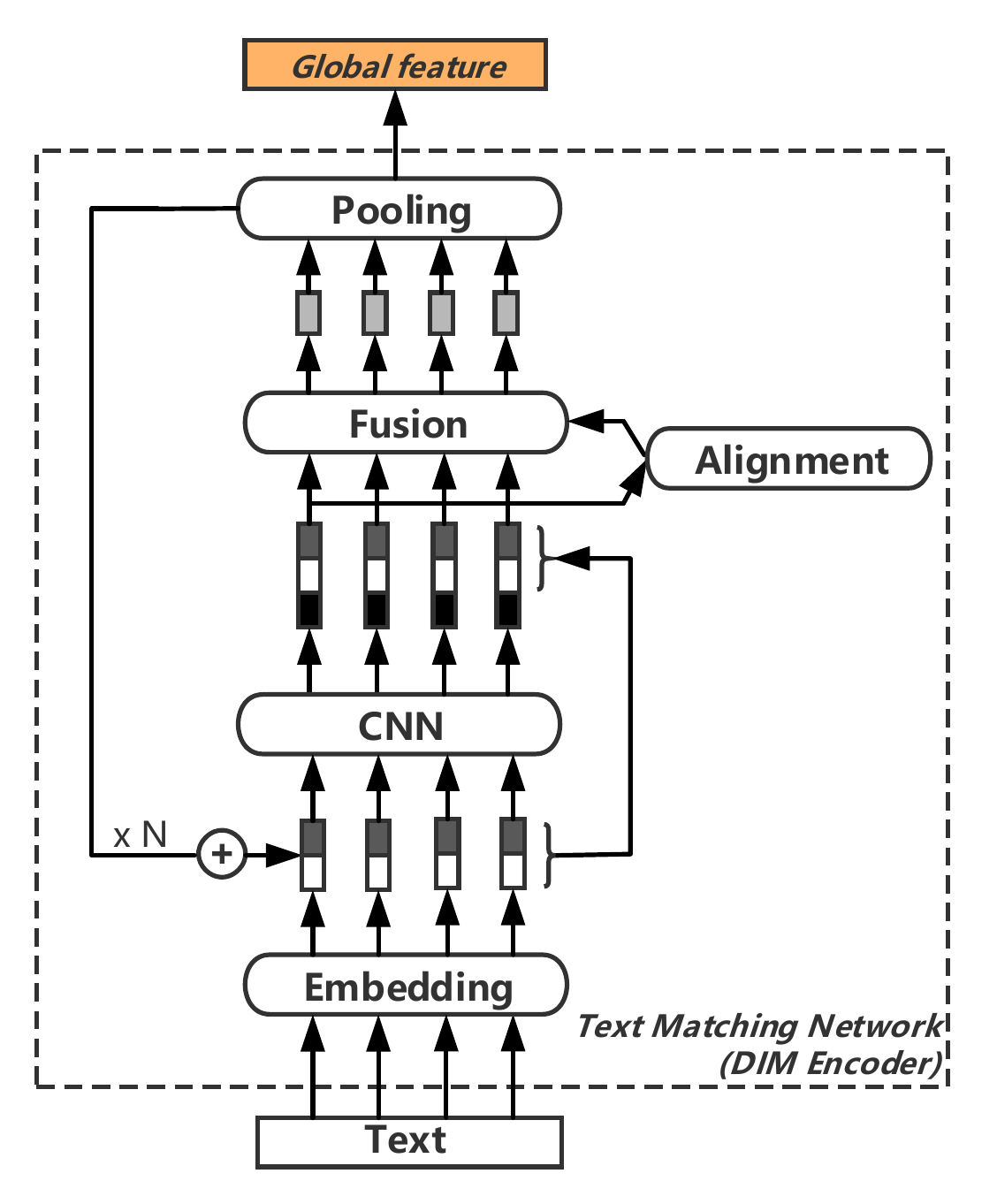}
  \caption{\textbf{Mutual Information Encoder.} 
  The baseline text matching neural architecture (RE2) is adopted as the DIM encoder in our model. Following settings in the original paper, the kernel size of the CNN layer is set to 3, and the number of CNN layer is tuned from 1 to 3. The block number $N$ of its augmented residual connections is tuned from 1 to 3. For experiments of SciTail, WikiQA, Quora and Harvard news, the high-level output is a 200 dimensional vector. For the experiment of SNLI, the output is a 150 dimensional vector.}
  \label{fig-TIM-RE2}
\end{figure}

For the global feature, we take the whole text matching neural network as the DIM encoder and use the its output as the high-level representation. We adopt RE2 as the base of the text matching network, which achieved the state-of-the-art on four well-studied datasets across three different text matching tasks. RE2 leverages the previous aligned features (Residual vectors), point-wise features (Embedding vectors), and contextual features (Encoded vectors), to maintain useful information of texts in one text matching task when information passes through its network. The detailed architecture of RE2 is illustrated in Figure \ref{fig-TIM-RE2}. An embedding layer firstly embeds discrete words. Three layers following the embedding layer are layers of encoding (CNN), alignment and fusion, which then process the sequences consecutively. The three layers are treated as one block in RE2. $N$ blocks are connected by an augmented version of residual connections. In the end, a pooling layer aggregates sequential representations into final vectors. More details can be referred in the original literature. 

As the high-level global feature output of the DIM encoder, the final vectors are then passed into the DIM discriminator network and trained. Simultaneously, the final vectors are also passed to and processed by a prediction layer to give the final prediction of text matching. We keep RE2's original network architecture (state of the art), then add DIM network on the base text matching network to help maximize useful information in the output representations used in the last step of matching prediction, which improves the performance to the text matching tasks and ensure the contrast experiments are reasonable.

\subsection{MI Maximization for Text Matching}
\label{sec:dim}
In our model, we define MI estimator and employ a discriminator to optimize the output representation ($E_\psi(X)$) of the input text data ($X$) by simultaneously estimating and maximizing MI, $\mathcal{I}(X; E_\psi$(X)), in both sides of the comparison.

\textbf{DIM Estimator.} To estimate MI, an appropriate lower-bound for the KL-divergence is necessary. Before DIM, MINE proposed a lower-bound to the MI based on the Donsker-Varadhan representation (DV, Donsker \& Varadhan, 1983) of the KL-divergence, shown as the following form:
\begin{align}
   \mathcal{I}(X; Y) & := \mathcal{D}_{KL}(\mathbb{J} || \mathbb{M})  \nonumber\\  
   & \geq \widehat{\mathcal{I}}^{(DV)}_{\omega}(X; Y) := \mathbb{E}_\mathbb{J}[T_{\omega}(x,y)] - \log \mathbb{E}_\mathbb{M}[e^{T_{\omega}(x,y)}],
    \label{eq:mine}
\end{align}
where $T_{\omega}: X\times{Y} \rightarrow \mathbb{R}$ is a discriminator function modeled by a neural network with parameters ${\omega}$. Based on the MINE estimator and the DIM local framework, we present our DIM estimator, maximizing the average estimated MI for text data and optimizing this local objective, described as following:
\begin{align}
    (\hat{\omega}, \hat{\psi})_L 
    &= {arg\,max}_{\omega, \psi} \frac{1}{M} \sum_{i=1}^{M} \widehat{\mathcal{I}}_{\omega, \psi}(C^{(i)}(X); E_\psi(X)),
    \label{eq:dim-estimator}
\end{align}
where $C(X)$ denotes local features converted from input texts by feature extraction. $E_{\psi}(X)$ is the learned high-level representation output of the pooling layer of the base text matching neural network RE2 with parameters $\psi$. The $\omega$ denotes the parameters of a DIM discriminator function modeled by a neural network. The subscript $L$ denotes ``local" for the DIM local framework. With our estimator, we next describe its DIM discriminator for MI maximization.

\begin{figure}[H]
    \centering
    \includegraphics[width=85mm]{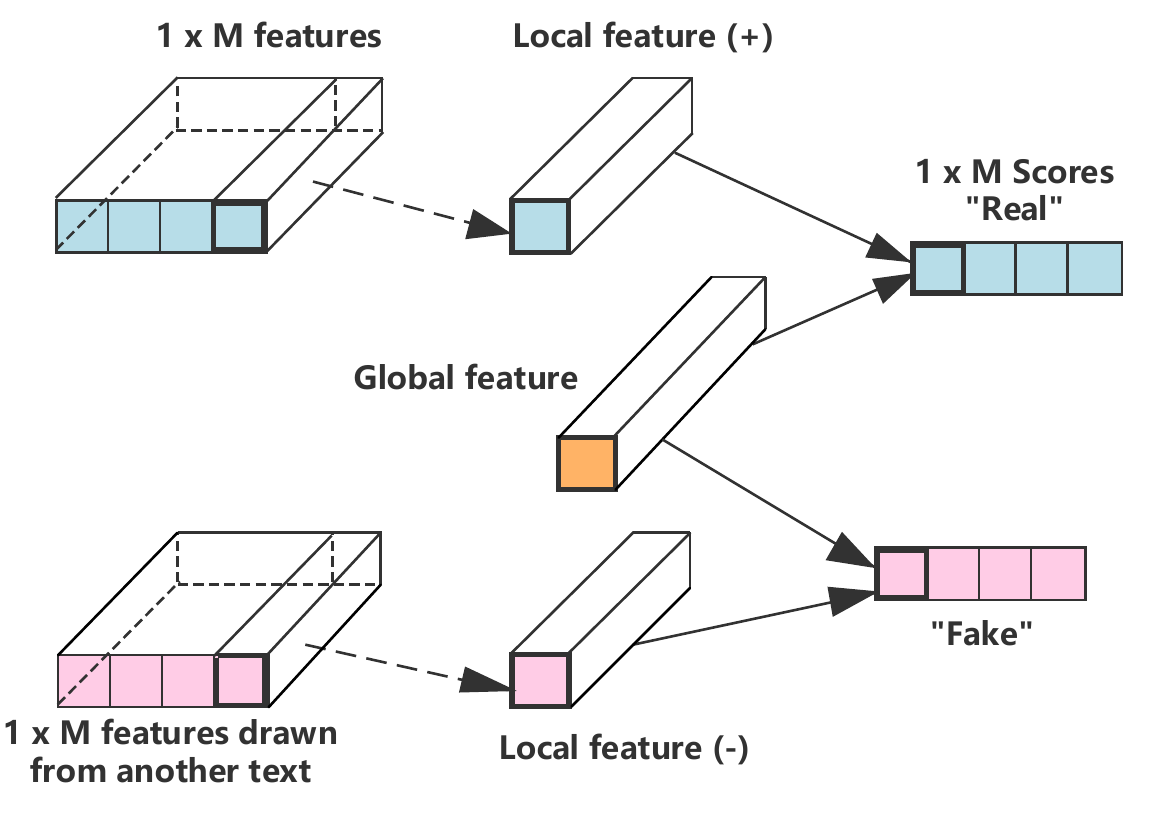}
    \caption{\textbf{Mutual Information Discriminator.} The global feature vector is concatenated with the lower-level feature map at every location. A $1\times{1}$ convolutional discriminator is used to score the `real' feature map vector pair, while the `fake' pair is produced by pairing the feature vector with a feature map from another text.}
    \label{fig-TIM-feature-map}
\end{figure}

\textbf{DIM Discriminator.} With the high-level output from the text matching network and the feature maps extracted from the same input text, we then concatenate this global feature vector with its relative lower-level feature maps at every location, $\{[C^{(i)}_{\psi}(x), E\psi(x)]\}_{i=1}^{1{\times}M}$, with $C(X)$ flattened in advance. Then our discriminator is formulated as:
  \begin{equation}\label{equal_local_discriminator}
    T^{(i)}_{\psi, \omega}(x, E_{\psi}(x)) = D_{\omega}([C^{(i)}(x), E_{\psi}(x)]),
\end{equation}
while fake feature maps are generated by combining global feature vectors with local feature maps coming from different texts, $x'$:
\begin{align}
    T^{(i)}_{\psi, \omega}(x', E_{\psi}(x)) = D_{\omega}([C^{(i)}(x'), E_{\psi}(x)]).
\end{align}
With the `real' and the `fake' feature maps, we introduce the local DIM concat-and-convolve network architecture ($D_{\omega}$), a $1\times1$ convnet with two 512-unit hidden layers, as DIM discriminator for our text matching model. The process is shown in Figure \ref{fig-TIM-feature-map}. Then `real' feature map and the `fake' feature map pass through the discriminators and get the $1{\times}M$ scores. The loss of MI for the input source text $t_s$ and target text $t_t$ of a text matching task can be calculated by: $\mathcal{L_M}$ = $\mathcal{L}_{t_s}$ + $\mathcal{L}_{t_t}$. The overall loss function can be defined as: $\mathcal{L}_{all}$ = $\mathcal{L_M}$ + $\mathcal{L_T}$, where $\mathcal{L_T}$ is the loss calculated by the base text matching neural network.

\section{Experiments}

\subsection{Experimental Setup}
\subsubsection{Benchmarks and Metrics}
We evaluated our proposed TIM-W and TIM-S model on four well-studied NLP tasks and a news dataset, as follows:\\
\textbf{Natural Language Inference.}
Stanford Natural Language Inference\footnote{\url{https://nlp.stanford.edu/projects/snli}} (SNLI) is a benchmark dataset for
natural language inference. In this task, the two input sentences are asymmetrical, one as ``premise'' and the other as ``hypothesis''.  We follow the setup of SNLI's original introduction
in training and testing. Accuracy is used as the evaluation metric for this dataset.\\
\textbf{Science Entailment.}
SciTail\footnote{\url{http://data.allenai.org/scitail}}
is an entailment classification dataset constructed
from science questions and answers. This dataset contains only two types of labels, entailment and neutral. We use the original dataset partition. It contains 27k examples in total. 10k examples are with entailment labels and the remaining 17k are labeled as neutral. Accuracy is used as the evaluation metric for this dataset\\
\textbf{Paraphrase Identification.}
This task is to decide whether one question is a paraphrase of the other between pairs of texts. We use the Quora dataset with 400k question pairs collected from the Quora website. The partition of dataset is the same as the one in \cite{wang2017bilateral}.And accuracy is used as the evaluation metric.\\
\textbf{Question Answering.}
For this task, we employ the WikiQA dataset\footnote{\url{https://www.microsoft.com/en-us/research/publication/wikiqa-a-challenge-dataset-for-open-domain-question-answering}}, which is a retrieval-based question answering dataset based on Wikipedia
.It contains questions and their candidate answers,with binary labels indicating whether a candidate sentence is a correct answer to the question it belongs to. Mean average precision (MAP) and mean reciprocal rank (MRR) are used as the evaluation metrics for this task.\\
\textbf{News Articles Title Content Match.}
We employ the Harvard news dataset, News Articles\footnote{\url{https://dataverse.harvard.edu/dataset.xhtml?persistentId=doi:10.7910/DVN/GMFCTR}}, for matching task. It contains news articles and we separate the title of each article from its content and do the data augmentation by randomly combining pairs of a title and content of an article. Most contents of the news articles have 1000 to 5000 words. We report matching accuracy. 
\subsubsection{Baselines and Implementations}
We implement our model based on \cite{yang2019simple} but train it on Nvidia 1080ti GPUs. Sentences in the dataset are all tokenized and converted to lower cases. We also perform a filter on meaningless symbols or emojis before embedding.The maximum sequence length is not limited. Word embeddings are initialized with 840B-300d GloVe word vectors (\cite{pennington2014glove}) and fixed during training process.

\subsection{Experimental Results}
The experimental results are described below:
\begin{itemize}
\item \textbf{Natural Language Inference.} Results on SNLI are shown in the first column of Table \ref{tab:experiment-result}. The performance of previous methods are quite close and we slightly outperform the state-of-the-art. Our method can perform well in the language inference task without any tasks-pecific modifications.

\item \textbf{Science Entailment.} Results on SciTail dataset are shown in the second column of Table \ref{tab:experiment-result}. Our method successfully improves the baseline model by 0.8\% and achieves a result 0.1\% over the state-of-the-art, which indicates our method is highly effective on this task.

\item \textbf{Paraphrase Identification.} Results on Quora are shown in third column of Table \ref{tab:experiment-result}. Our method also lifts the accuracy of baseline model by 0.4\% and achieves higher results than all previous methods.

\item \textbf{Question Answering.} Results on WikiQA are shown in the last column of Table \ref{tab:experiment-result}. Small improvements are made by our methods on this IR task, which indicates our method also fits IR tasks well.

\item \textbf{News Article Title Content Match.} Results on harvard news dataset are shown in Table \ref{tab:news-result}.
\end{itemize}

\begin{table*}
    \begin{tabular}{ |p{2.66cm}|p{0.52cm}||p{2.66cm}|p{0.52cm}||p{2.76cm}|p{0.52cm}||p{2.73cm}|p{0.74cm}|p{0.74cm}|  }
        \hline
        \multicolumn{2}{|c||}{\textbf{SNLI}} & 
        \multicolumn{2}{c||}{\textbf{SciTail}} &
        \multicolumn{2}{c||}{\textbf{Quora}} &
        \multicolumn{3}{c|}{\textbf{WikiQA}} \\ 
        \hline
        \textbf{Model} & \textbf{Acc.} & \textbf{Model} & \textbf{Acc.} & \textbf{Model} & \textbf{Acc.} & \textbf{Model} & \textbf{MAP} & \textbf{MRR} \\
        \hline
        BiMPM \newline \cite{wang2017bilateral} & 86.9 & ESIM \newline \cite{chen2017enhanced} & 70.6 & BiMPM \newline \cite{wang2017bilateral} & 88.2 & ABCNN \newline \cite{yin2016abcnn} & 0.6921 & 0.7108 \\ \hline
        ESIM \newline \cite{chen2017enhanced} & 88.0 & DecAtt \cite{parikh2016decomposable} & 72.3 & pt-DecAttn-word \newline \cite{tomar2017neural} & 87.5 & KVMN \newline \cite{miller2016key} & 0.7069 & 0.7265 \\ \hline
        MwAN \newline \cite{tan2018multiway} & 88.3 & DGEM \newline \cite{Khot2018SciTaiLAT} & 77.3 & pt-DecAttn-char \newline \cite{tomar2017neural} & 88.4 & BiMPM \newline \cite{wang2017bilateral} & 0.718 & 0.731 \\ \hline
        CAFE \newline \cite{tay2018compare} & 88.5 & HCRN \newline \cite{tay2018hermitian} & 80.0 & MwAN \newline \cite{tan2018multiway} & {89.1}  & IWAN \newline \cite{shen2017inter} & 0.733 & 0.750 \\ \hline
        SAN \newline \cite{liu2018stochastic} & 88.6 & CAFE \newline \cite{tay2018compare} & 83.3 & CSRAN \newline \cite{tay2018co} & {\bf 89.2} & CA \cite{wang2017compare} & {0.7433} & {0.7545}\\ \hline
        CSRAN \newline \cite{tay2018co} & 88.7 & CSRAN \newline \cite{tay2018co} & {\bf 86.7} & SAN \newline \cite{liu2018stochastic} & {\bf 89.4} & HCRN \newline \cite{tay2018hermitian} & {0.743} & {0.756} \\ \hline
        RE2 \newline \cite{yang2019simple} & {\bf 88.9} & RE2 \newline \cite{yang2019simple} & {\bf 86.0} & RE2 \newline \cite{yang2019simple} & {\bf 89.2} & RE2 \newline \cite{yang2019simple} & {\bf 0.7452} & {\bf 0.7618} 
        \\\hline
        \hline
        TIM-W (ours) & {\bf 88.9} & TIM-W (ours) & {\bf 86.8} & TIM-W (ours)  & {\bf 89.6} & TIM-W (ours) & {\bf 0.7516} & {\bf 0.7685}\\
         & {\bf$\pm$0.1} & & {\bf$\pm$0.1} & & {\bf$\pm$0.3} & & {$\pm$0.02} & {$\pm$0.02}\\
        
        TIM-S (ours) & 88.3 & TIM-S (ours) & 86.2 & TIM-S (ours) & 87.8 & TIM-S (ours) & {0.7181} & {0.7387}\\
         & $\pm$0.1 & & $\pm$0.1 & & $\pm$0.5 & & {$\pm$0.02} & {$\pm$0.02} \\
        \hline
    \end{tabular}
    \caption{Experimental results on four datasets: SNLI, SciTail, Quora and WikiQA.}
    \label{tab:experiment-result}
\end{table*}

\begin{table}
    \centering
    \small
    \begin{tabular}{|l|l|}
    \hline
    {\bf Model} & {\bf Acc(\%)}\\
    \hline
    RE2 \cite{yang2019simple} & {93.18} \\
    \hline
    TIM-S (ours): D=12, M=10 & {\bf 96.59}\\
    TIM-S (ours): D=20, M=10 & {95.83}\\
    TIM-S (ours): D=20, M=20 & {95.45}\\
    TIM-S (ours): D=6, M=10 & {95.11}\\
    TIM-S (ours): D=6, M=5 & {94.70}\\
    TIM-W (ours) & {94.14}\\
    \hline
    \end{tabular}
    \caption{Experimental results on Harvard news dataset, with the infulence of $M$ and $D$ in TIM-S mode.}
    \label{tab:news-result}
\end{table}

In all, our proposed method achieves equal or even better performance on par with the state-of-the-art on four well-studied datasets across three different tasks.\\
\textbf{Analysis of Results. } TIM-W mode on SNLI, Quora, Scitail and WikiQA achieves better accuracy for feature extraction on the word level because of short texts. For feature extraction on the segment level for longer texts, TIM-S mode suits better according to the experiments on the News Article dataset. And without introducing high-dimension pretrained word embedding, TIM-S is significantly faster on the long texts than TIM-W. 
\\
\textbf{Influence of $M$ and $D$. } $M$ is the shape size of the local feature in both TIM-W and TIM-S, and $D$ is the segment size only need to be set in TIM-S. First, for the TIM-W used in SNLI, Quora, Scitail and WikiQA, we tune $M$ from 1 to 3. Texts in the four datasets are relatively short and $M$ should not be greater than the word number of the short text. Otherwise a short text will only be converted to just one feature map, which will cause loss of structural information in the text. Second, in the experiments under TIM-S mode for the content field in the News dataset, we tune the segment size (words, $D$) and the shape ($1{\times}M$) of the fixed-size feature maps. We set $D=12$ and $M$=$10$, which means each local feature map contains $10$ segments and each segment has $12$ word indexes. For the setting of $M$ and $D$, when we enlarge both $D$ and $M$, each feature map block will have more zeros padded so that it becomes more difficult to maximize the useful local information from sparse feature maps. But when both $D$ and $M$ are set to be small, the TIM-S mode actually becomes TIM-W mode, which is not suitable for long texts. This means when the shape of the feature map in TIM-S becomes smaller, more local structure information is lost in the MI maximization process. The influence of $M$ and $D$ is illustrated in Figure \ref{tab:news-result}.\\
\textbf{Case Study. } Aligning tokens between two texts is a key stage of the baseline model and achieves remarkable improvements on text matching. But incorrect concentration on the text positions during finite number of alignment operations (3 times), may result in failure of predictions. For example, in a pair from WikiQA, ``who is basketball star \textit{antoine walker}" and ``\textit{Antoine Devon Walker} (born August 12, 1976) is an American former professional basketball player", there is a middle name in the player's name. And in another pair, ``what day is \textit{st. patricks} day" and ``\textit{Saint Patrick}'s Day or the Feast of \textit{Saint Patrick} (the Day of the Festival of \textit{Patrick}) is a cultural and religious holiday celebrated on 17 March", the person's name appears at multiple positions in one text. Compared to the baseline, MI maximization with powerful neural networks helps to model local semantics and improve text matching predictions more efficiently. Our model gets better prediction results on these cases. Meanwhile, for richer features can bring better MI estimation results, we will investigate better feature extraction methods with MI neural estimation for NLP tasks in future work. 

\section{Conclusions}
In this paper, we propose a new neural architecture with deep mutual information estimation to learn more effective and high-quality text representations in text matching tasks. By maximizing the mutual information between each input and output pairs, our method retains more useful information in the learned high-level representations. Moreover, we split text into segments and treat these segments as local features. This helps preserve the complex structural information and solve the structure difficulty in text matching on varying-length texts. Then we leverage local mutual information maximization method to solve the information loss problem from complex text structures in text matching frameworks. The experiment results on various text matching tasks also demonstrate the effectiveness of our model.

\newpage

\bibliographystyle{named}
\bibliography{main}

\end{document}